\documentclass[10pt,twocolumn,letterpaper]{article}

\usepackage[pagenumbers]{wacv} 

\usepackage{graphicx}
\usepackage{amsmath}
\usepackage{amssymb}
\usepackage{booktabs}
\usepackage{multirow}
\usepackage{caption}
\usepackage{booktabs}
\usepackage{subcaption}

\usepackage{times}
\usepackage{epsfig}
\usepackage{amsmath}
\usepackage{amssymb}
\usepackage{booktabs}
\usepackage{multirow}
\usepackage{cite}
\usepackage[pagebackref,breaklinks,colorlinks]{hyperref}

\usepackage[capitalize]{cleveref}
\crefname{section}{Sec.}{Secs.}
\Crefname{section}{Section}{Sections}
\Crefname{table}{Table}{Tables}
\crefname{table}{Tab.}{Tabs.}

\makeatletter
\newcommand\notsotiny{\@setfontsize\notsotiny\@vipt\@viipt}
\makeatother
\graphicspath{{./Images/}}

\usepackage[capitalize]{cleveref}
\crefname{section}{Sec.}{Secs.}
\Crefname{section}{Section}{Sections}
\Crefname{table}{Table}{Tables}
\crefname{table}{Tab.}{Tabs.}

\def\wacvPaperID{656} 
\def\confName{WACV}
\def\confYear{2024}

\begin{document}
	
	\title{Context-based Interpretable Spatio-Temporal Graph Convolutional Network for Human Motion Forecasting}
	
	\author{Edgar Medina, Leyong Loh, Namrata Gurung, Kyung Hun Oh, Niels Heller, \\
		QualityMinds GmbH\\
		{\tt\small \{edgar.medina, leyong.loh, namrata.gurung, kyung-hun.oh, niels.heller\}@qualityminds.de}
}

\maketitle

\begin{abstract}
	Human motion prediction is still an open problem extremely important for autonomous driving and safety applications. Due to the complex spatiotemporal relation of motion sequences, this remains a challenging problem not only for movement prediction but also to perform a preliminary interpretation of the joint connections. In this work, we present a Context-based Interpretable Spatio-Temporal Graph Convolutional Network (CIST-GCN), as an efficient 3D human pose forecasting model based on GCNs that encompasses specific layers, aiding model interpretability and providing information that might be useful when analyzing motion distribution and body behavior. Our architecture extracts meaningful information from pose sequences, aggregates displacements and accelerations into the input model, and finally predicts the output displacements. Extensive experiments on Human 3.6M, AMASS, 3DPW, and ExPI datasets demonstrate that CIST-GCN outperforms previous methods in human motion prediction and robustness. Since the idea of enhancing interpretability for motion prediction has its merits, we showcase experiments towards it and provide preliminary evaluations of such insights here. \footnote{available code: \href{https://github.com/QualityMinds/cistgcn}{https://github.com/QualityMinds/cistgcn}}
	
\end{abstract}

\vspace{-2mm}
\section{Introduction}
\vspace{-0.5mm}
\label{sec:intro}


\begin{figure*}
	\centering
	\includegraphics[width=0.75\linewidth]{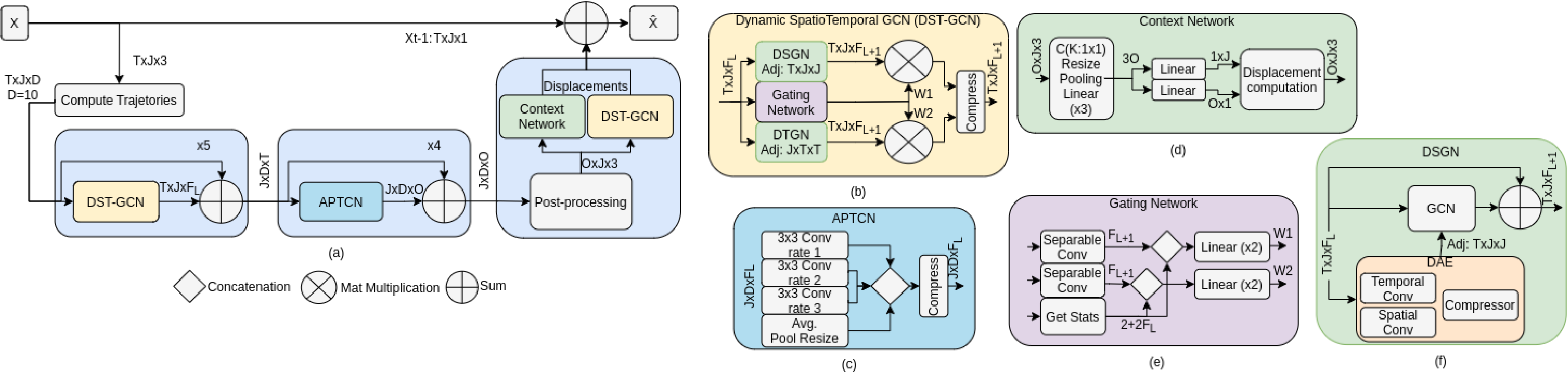}
	\begin{subfigure}{0\textwidth}
		\phantomcaption
		\label{fig:architecture}
	\end{subfigure}
	\begin{subfigure}{0\textwidth}
		\phantomcaption
		\label{fig:DSTGCN}
	\end{subfigure}
	\begin{subfigure}{0\textwidth}
		\phantomcaption
		\label{fig:APTCN}
	\end{subfigure}
	\begin{subfigure}{0\textwidth}
		\phantomcaption
		\label{fig:Context}
	\end{subfigure}
	\begin{subfigure}{0\textwidth}
		\phantomcaption
		\label{fig:Gating}
	\end{subfigure}
	\begin{subfigure}{0\textwidth}
		\phantomcaption
		\label{fig:DAE}
	\end{subfigure}
	
	\vspace{-2.5mm}
	\caption{Illustration of our method. (a) Overview of the proposed CIST-GCN. $X$ and $\hat{X}$ are the input and output respectively. (b) The basic block of DST-GCN, (c) the Atraus Pyramid TCN, and (d) Context Network. More detailed, (e) Gating network weights the output of DSGN and DTGN, and (f) Dynamic Adjacency Encoder (DAE) to compute the adjacency matrices.}
	\label{fig:overview}
\end{figure*}

Human motion prediction plays a critical role in autonomous driving, robotics, and safety applications. In the recent past, several methods for human motion prediction and modeling have led to significant results with the use of neural networks \cite{Lyu2022d}. Recently, the main approaches to tackle this task have been by means of Graph Convolutional Networks (GCN) \cite{Mao2019, Li2020, Mao2020, Li2021, Yan2021, Zhou2021e, Sofianos2021b, Zhong2022, Ma2022}, Recurrent Networks (RNN) \cite{Martinez2017a, Guo2019, Mao2019, Wang2021, Liu2021, Su2022} and GANs \cite{Lyu2022d}. Although in the last years, RNN-based models were the most effective methods, they come with the drawback of vanishing or exploding gradients. Recent approaches mix more sophisticated architectures such as Gated recurrent units (GRU) \cite{Su2022} or transformers \cite{Aksan2020, Lyu2022d} with feature extraction using CNN or a gate system in the hidden states. Alternately, the GAN-based approach \cite{Lyu2022d} is another methodology for generating the sequence from a hidden vector, but this approach neglects the kinematic dependencies between pose joints and ignores the temporal correlation between frames. Instead, GCNs have received increasing attention because this architecture can find a temporal relation between poses and can understand relationships among joints.

A second branch in this work uses interpretability concepts, initial stages were inspired by class-activation \cite{Zhou2015} or saliency maps. However,  the authors using saliency maps only performed a visual analysis without providing formal statistical evaluations. \cite{Zhou2015, Zhang2020a, Liu2023a}. Nowadays, more sophisticated methods quantify the error or even measure the uncertainty level of movement predictions \cite{Mukhoti2021, Moreno-Vera2023}. Despite the great advances in interpretability of CNNs in classification tasks, GCNs are not yet properly covered \cite{Kazi2021, Cui2022}, especially for regression tasks (such as motion prediction) and not classification tasks.

The motivation for designing this architecture is to close the gap between motion prediction and interpretability and apply it to real-world problems to gain meaningful insights into why the model predicted a specific output. In our proposed Context-based Interpretable Spatio-Temporal GCN (CIST-GCN) architecture, we embed GCN layers which provide sample-specific adjacency matrices and importance vectors to explain motion forecasting. The matrices are composed of learnable parameters while the importance vectors are generated at output layers a mix of CNN and MLP layers. It stands to reason that these features are human-interpretable. While we provide some of such interpretations in this paper. To the best of our knowledge, this is the first work that drives a GCN-based architecture in this direction. Finally, data augmentation has been added to speed up the training and also make the system more robust to possible data glitches that may occur in production use-case such as faulty 3D-transformations or falsy reconstructed 3d poses. We conducted experiments to study the robustness of our model against out-of-distribution (OOD) data samples, for example, rotations, glitches in poses. 

Our approach consistently obtains comparable results to the previous results on short- and long-term motion prediction by training a single unified model for both settings. Specifically, we achieve superior performance in 6 out of 15 actions on the Human3.6M benchmark, while remaining comparable in the other motion predictions. Our model also surpasses previous works in 3DPW and 12 out of 16 actions on ExPI datasets, while also achieving comparable results on the AMASS benchmark. The main contributions of our work can be summarized as follows: 1) we propose a new architecture that provides not only human motion prediction, but also interpretability to some extent given an input sample, 2) we perform extensive experiments on Human3.6M \cite{Ionescu2014}, AMASS \cite{Mahmood2019}, 3DPW \cite{VonMarcard2018}, and ExPI \cite{Guo2021} datasets, showing that quantitative and qualitative results are comparable to state-of-the-art (SOTA) models, 3) we discuss the different extents of interpretations such as relation matrices, and importance vectors, and 4) we perform robustness experiments that showcases our model to be better than existing SOTA models for ODD samples.

\section{Related work}
\label{sec:relat}
\subsection{Motion Prediction}
Initial research prove the strength of relations between joint connections in pose sequences both in the temporal and spatial domains \cite{Lyu2022d}. Subsequent work \cite{Mao2019, Li2021, Liu2021b, Yan2021, Zhou2021e} research deeper into this approach by grouping the input joints in several ways or merging GCN, CNN, and GRU layers to learn the graph connectivity (forming a spatio-temporal graph). In \cite{Sofianos2021b}, STS-GCN receives the 3D coordinates as input but uses two GCNs to encode sequentially temporal and spatial data in every encoding layer to feed the decoder. This decoder is composed of a 4-block Temporal Convolution Network (TCN) \cite{Mao2020} that converts input frames into output frames. Overall, this work requires a lower number of parameters than previous approaches and inspired recent architectures.

Another newest branch, such as MotionMixer \cite{Bouazizi2022a}, employs linear and feature mix layers to merge information. Although the results are promising, there are problems in understanding which joints or frames may be relevant for further analysis such as prediction reliability or action clustering. Since the architecture behaves as a black box model, we cannot obtain correlation or relationship matrices coming from the model. Also, experiments such as the application of 3D transformations can illustrate its limitations.

In \cite{Fu2022b}, several modifications to DSTD-GCN are proposed such as dynamic spatial and temporal graph convolutions are presented as separate units, allowing features to be learned independently. The authors suggest using constrained training with different strategies. Later, they show that relations can be acquired in unshared sample-specific forms, reducing MPJPE significantly. The impact of this approach on metrics inspired us to incorporate learnable adjacency matrices in all our GCN layers, removing the need for duplicating dynamic spatial GCN. This results in fewer parameters compared to DSTD-GCN. Also, in Section \ref{sec:discussion}, we explore the potential application of this as an interpretable output for individual samples. In \cite{Zhong2022}, a gating network is proposed to generate blending coefficients that weighs the most meaningful features of the adjacency matrices from GCNs. Number of weight vectors for the temporal and spatial GCNs are equal. Also, the authors suggest that these acquired vectors could aid in action grouping and emphasize the most important features in motion prediction. Motivated by this approach, we use weighting vectors in a similar manner, adding layers with interpretable variables, but reducing the number of channels in every layer required for motion prediction. We show how similar movements have similar interpretable patterns in later sections.

\subsection{Model Interpretability}
The application of the CAM \cite{Zhou2015, Zhang2020a, Qin2021} methods is limited to GNN structures due to they have special requirements and assume heuristically that the final node embedding can reflect the input importance. This assumption may be wrong \cite{Yuan2020}. While saliency map techniques are commonly employed for model interpretability, they are primarily designed for images. Many of these methods were evaluated solely in classification tasks, applied as independent post-training steps, or require subsequent visual validation. Given the nature of our problem, graph interpretation is required \cite{Yuan2020}. In \cite{Kazi2021}, a model capable of incorporating interpretable attention is proposed. Later, applications started to use interpretable GCNs (IGCN) \cite{Kazi2018, Kazi2021, Cui2022}. Other approaches \cite{Li2022, Dai2021, Xie2022, Longa2022} need extra heavy computation to obtain interpretable features. A new method called GraphLIME \cite{Huang2022a} proposed a generic GNN-model explanation framework consisting of a local interpretable model explanation. Our model has similarities with IGCN and GraphLIME methods. However, we distinguish ourselves by explaining the relationship between frame-to-frame and joint-to-joint predictions, which deals with a subset of graph data. Specifically, GraphLIME \cite{Huang2022a} offers prediction explanations for GNN architectures, whereas our model provides interpretation output alongside motion predictions in a specific data structure. While IGCN \cite{Kazi2021} has shown combining prediction and interpretation on classification tasks with diverse situations, it only was tested on classification tasks and different data structures. In \cite{Dai2021}, self-explainable GNN can find K-nearest labeled nodes for the unlabeled nodes to explain the classification output. However, this approach only was tested with a synthetic dataset on classification tasks not related to motion movements. Also, it is well-known that post-hoc explanations can suffer bias and misrepresentation due to interpretations are not directly obtained from the GNNs \cite{Dai2021}. In contrast, our model predicts not only motion but also its interpretations by means of adjacency matrices and weighting vectors, solely utilizing GCN layers without requiring external post-processing methods while being applied to regression tasks.

\section{Methodology}
\label{sec:metho}

\subsection{Problem Formalization}
We define the body motion as a sequence of poses $X \in \mathbb{R}^{T \times J \times D}$ where $T$ and $J$ define a number of frames and joints from the sequence, and $D$ parameterizes each body joint dimension for angles or 3D coordinates. Our model receives a historical input poses $X_{in} = X_{0:t_1} = {x_0,x_1,...,x_{t_1-1}} \in \mathbb{R}^{t_1 \times J \times D}$ and predict a sequence of poses $\hat{X} = X_{t_1:t_2} = {\hat{x}_{t_1},\hat{x}_{t_1+1},...,\hat{x}_{t_1+t_2}} \in \mathbb{R}^{t_2 \times J \times D}$. The math representation for an adjacency matrix is given by $A \in \mathbb{R}^{P \times P}$ where $P \in \mathbb{R}^{n}$ is the node representation.

\subsection{Review of GCN}
\textbf{Graph Convolutional Networks}
Graph Convolutions (GCs) are suitable for non-grid data, where data is represented by a set of nodes (e.g. x,y,z coordinates) carrying n-dimensional information. When GCs are stacked sequentially then they together become a GCN. In this work, such a set of nodes is called a pose, and an adjacency matrix shows the connection between pairs of nodes from the whole graph. Formally, let $H^l \in \mathbb{R}^{P \times F^l}$, $A^l \in \mathbb{R}^{P \times P}$ be the input and the adjacency matrix at the current layer $l$, whereas the trainable parameters at current layer $l$ are represented by $W^l \in \mathbb{R}^{F^l \times F^{l+1}}$. $F$ is the number of channels of this layer. We show the mathematical operation in Eq. \ref{eq:gcn} where $\sigma$ is the activation function and $H^{l+1} \in \mathbb{R}^{P \times F^{l+1}}$ is the GC output.

\begin{equation}
	H^{l+1} = \sigma(A^l H^l W^l)
	 \label{eq:gcn}
\end{equation}

\textbf{Spatio-Temporal GCN}
Given that our problem contains a temporal factor in the data, we thus have a spatial-temporal graph. To process this graph we use two graph convolution operations, for the spatial and temporal domains, just like the STS-GCN  model \cite{Sofianos2021b}, which takes the interactions of the temporal evolution and the spatial joints. Spatial and temporal graph convolutions are presented in Eq. \ref{eq:stsgcn}. Where $W_s^l$ and $W_t^l$ $\in \mathbb{R}^{F^l \times F^{l+1}}$ are trainable parameters. Graph convolutions are separable if we operate independently and stack later, as argued later in \cite{Fu2022b}.

\begin{equation}
	H^{l+1} = \sigma(A_{t}^l A_{s}^l H^l W^l) = \sigma(A_t^l (A_s^l H^l W_s^l) W_t^l)
	 \label{eq:stsgcn}
\end{equation}

We perform a similar approach and demonstrate via experimentation this operation is stable in Eq. \ref{eq:spatial}. Where $W_D^l$ $\in \mathbb{R}^{F^l \times F^{l+1}}$ are trainable parameters, and $D$ represents the temporal or spatial domain.

\begin{equation}
	H_D^{l+1} = \sigma(A_D^l H_D^l W_D^l)
	\label{eq:spatial}
\end{equation}

\begin{table*}[ht]
	\begin{center}
		\notsotiny
		\caption{Performance comparison for motion prediction using MPJPE in every action from the Human3.6M dataset. ($^*$) implies metric is computed by us using our pipeline and the standard metric. ($^\dag$) metric takes the average MPJPE over all previous frames.}
		\begin{tabular}{c|c@{\hspace{5pt}}c@{\hspace{5pt}}c@{\hspace{5pt}}c@{\hspace{5pt}}c@{\hspace{5pt}}c@{\hspace{5pt}}|c@{\hspace{5pt}}c@{\hspace{5pt}}c@{\hspace{5pt}}c@{\hspace{5pt}}c@{\hspace{5pt}}c@{\hspace{5pt}}|c@{\hspace{5pt}}c@{\hspace{5pt}}c@{\hspace{5pt}}c@{\hspace{5pt}}c@{\hspace{5pt}}c@{\hspace{5pt}}|c@{\hspace{5pt}}c@{\hspace{5pt}}c@{\hspace{5pt}}c@{\hspace{5pt}}c@{\hspace{5pt}}c@{\hspace{5pt}}}
			\toprule 
			& 
			\multicolumn{6}{c|}{Walking} & 
			\multicolumn{6}{c|}{Eating} &
			\multicolumn{6}{c|}{Smoking}&
			\multicolumn{6}{c}{Discussion} \\
			
			Time (ms) &
			80 & 160 & 320 & 400 & 560 & 1000 &
			80 & 160 & 320 & 400 & 560 & 1000 &
			80 & 160 & 320 & 400 & 560 & 1000 &
			80 & 160 & 320 & 400 & 560 & 1000 \\
			\midrule
			
			ConvSeq2Seq\cite{Li2018} & 17.7 & 33.5 & 56.3 & 63.6 & 72.2 & 82.3 & 11.0 & 22.4 & 40.7 & 48.4 & 61.3 & 87.1 & 11.6 & 22.8 & 41.3 & 48.9 & 60.0 & 81.7 & 17.1 & 34.5 & 64.8 & 77.6 & 98.1 & 129.3 \\
			
			
			Traj-CNN\cite{Liu2021b} & 11.9 & 22.5 & 38.7 & 45.7 & 54.5 & 62. & 8.4 & 16.6 & 32.4 & 39.8 & 53.5 & 78.4 & 8.4 & 16.2 & 31.1 & 37.6 & 49.3 & 72.3 & 11.7 & 26.3 & 57.3 & 70.4 & 91.4 & 122.7 \\
			
			STS-GCN\cite{Sofianos2021b}$^*$ & 12.0 & 23.0 & 41.5 & 48.5 & 56.6 & 62.7 & 7.9 & 16.8 & 33.4 & 40.7 & 53.1 & 76.7 & 7.5 & 15.7 & 31.3 & 38.4 & 50.7 & 73.1 & 11.4 & 26.4 & 57.8 & 71.1 & 91.2 & 120.8 \\
			
			MSR-GCN\cite{Dang2021} & 12.2 & 22.6 & 38.6 & 45.2 & 52.7 & 63.0 & 8.4 & 17.0 & 33.0 & 40.4 & 52.5 & 77.1 & 8.0 & 16.3 & 31.3 & 38.2 & 49.4 & 71.6 & 12.0 & 26.8 & 57.1 & 69.7 & 88.6 & 117.6 \\
			
			MultiAttention\cite{Mao2021} & \textbf{9.9} & \textbf{19.3} & \textbf{33.7} & \textbf{39.0} & \textbf{46.2} & \textbf{57.1} & 7.9 & 17.5 & 37.4 & 45.2 & \textbf{48.6} & 73.7 & 7.0 & 14.3 & \textbf{25.4} & \textbf{29.0} & \textbf{46.5} & \underline{68.7} & \textbf{8.6} & \underline{22.8} & \textbf{51.0} & \textbf{64.0} & \textbf{85.2} & 117.5 \\
			
			DSTD-GCN\cite{Fu2022b} & 11.0 & 22.4 & 38.8 & 45.2 & 52.7 & 59.8 & 7.0 & 15.5 & 31.7 & 39.2 & 51.9 & 76.2 & \textbf{6.6} & 14.8 & 29.8 & 36.7 & 48.1 & 71.2 & \underline{10.0} & 24.4 & 54.5 & 67.4 & 87.0 & 116.3 \\
			
			MotionMixer\cite{Bouazizi2022a} & 10.8 & 22.4 & 36.5 & 42.4 & - & 59.9 & 7.7 & \textbf{14.0} & 27.3 & \textbf{36.1} & - & 76.6 & \underline{7.1} & \textbf{14.0} & 29.1 & 36.8 & - & \textbf{68.5} & 10.2 & \textbf{22.5} & \textbf{51.0} & \underline{64.1} & - & 117.4 \\
			
			PGBIG \cite{Ma2022} & \underline{10.2} & \underline{19.8} & \underline{34.5} & \underline{40.3} & \underline{48.1} & \underline{56.4} & 7.0 & \underline{15.1} & \underline{30.6} & 38.1 & 51.1 & 76.0 & \textbf{6.6} & \underline{14.1} & \underline{28.2} & \underline{34.7} & \textbf{46.5} & 69.5 & \underline{10.0} & 23.8 & 53.6 & 66.7 & 87.1 & 118.2 \\
			
			

			\midrule
			M16 &   
			12.0 & 23.6 & 41.0 & 46.8 & 54.3 & 61.9 &
			\underline{6.9} & \underline{15.1} & \underline{30.6} & 37.8 & 50.8 & \underline{75.3} &
			7.5 & 15.7 & 31.5 & 38.4 & 49.7 & 71.5 &
			10.3 & 24.1 & 52.8 & 65.8 & \underline{85.9} & \textbf{115.1} \\
			
			M32 & 
			11.8 & 23.4 & 40.5 & 46.5 & 54.1 & 61.3 & 
			\textbf{6.7} & 14.8 & \textbf{29.8} & \underline{36.8} & \underline{49.8} & \textbf{74.7} &
			7.3 & 15.6 & 31.0 & 38.0 & \underline{49.4} & 70.7 & 
			10.2 & 23.7 & \underline{52.3} & 65.3 & 86.1 & \underline{115.9} \\
			
			\midrule 
			& 
			\multicolumn{6}{c|}{Directions} & 
			\multicolumn{6}{c|}{Greeting} &
			\multicolumn{6}{c|}{Phoning}&
			\multicolumn{6}{c}{Posing} \\
			
			Time (ms) &
			80 & 160 & 320 & 400 & 560 & 1000 &
			80 & 160 & 320 & 400 & 560 & 1000 &
			80 & 160 & 320 & 400 & 560 & 1000 &
			80 & 160 & 320 & 400 & 560 & 1000 \\
			\midrule 
			
			
			ConvSeq2Seq\cite{Li2018} & 13.5 & 29.0 & 57.6 & 69.7 & 86.6 & 115.8 & 22.0 & 45.0 & 82.0 & 96.0 & 116.9 & 147.3 & 13.5 & 26.6 & 49.9 & 59.9 & 77.1 & 114.0 & 16.9 & 36.7 & 75.7 & 92.9 & 122.5 & 187.4 \\
			
			
			Traj-CNN\cite{Liu2021b} & 8.7 & 19.3 & 43.6 & 54.4 & 74.6 & 109.4 & 15.8 & 35.1 & 73.6 & 88.9 & 110.8 & 149.6 & 10.1 & 20.5 & 42.0 & 51.9 & 69.3 & 104.4 & 12.1 & 26.9 & 62.4 & 79.3 & 108.4 & 170.9 \\
			
			STS-GCN\cite{Sofianos2021b}$^*$ & 7.8 & 18.7 & 42.6 & 53.2 & 71.0 & 102.1 & 15.3 & 35.0 & 73.4 & 89.1 & 112.2 & 143.9 & 9.5 & 20.4 & 41.6 & 51.1 & 68.3 & 103.7 & 11.6 & 27.6 & 63.8 & 81.2 & 111.7 & 168.4 \\
			
			MSR-GCN\cite{Dang2021} & 8.6 & 19.6 & 43.3 & 53.8 & 71.2 & 100.6 & 16.5 & 37.0 & 77.3 & 93.4 & 116.3 & 147.3 & 10.1 & 20.7 & 41.5 & 51.3 & 68.3 & 104.3 & 12.8 & 29.4 & 67.0 & 85.0 & 116.3 & 174.3 \\
			
			MultiAttention\cite{Mao2021} & 11.3 & 22.9 & 50.6 & 62.6 & 72.4 & 105.7 & \underline{12.9} & 26.6 & 68.2 & 85.4 & \textbf{100.5} & 136.7 & 11.2 & 19.6 & \underline{37.7} & \textbf{44.1} & \underline{66.5} & 104.6 & \underline{9.8} & \underline{23.7} & 62.2 & 78.7 & \textbf{105.8} & 172.9 \\
			
			DSTD-GCN\cite{Fu2022b} & \textbf{6.9} & \textbf{17.4} & \underline{41.0} & \underline{51.7} & \textbf{69.0} & \textbf{99.0} & 14.3 & 33.5 & 72.2 & 87.3 & 108.7 & 142.3 & \underline{8.5} & 19.2 & 40.3 & 49.9 & 66.7 & \textbf{102.2} & 10.1 & 25.4 & 60.6 & 77.3 & 106.5 & \textbf{163.3} \\
			
			MotionMixer\cite{Bouazizi2022a} & 8.3 & 18.1 & 43.8 & 53.4 & - & 105.4 & \textbf{12.8} & 33.4 & \textbf{62.3} & 82.2 & - & 136.5 & 10.0 & 20.1 & \textbf{37.4} & 51.1 & - & 104.4 & 11.7 & \textbf{23.3} & 62.4 & 79.5 & - & 174.9 \\
			
			PGBIG \cite{Ma2022} & \underline{7.2} & \underline{17.6} & \textbf{40.9} & \textbf{51.5} & \underline{69.3} & \underline{100.4} & 15.2 & 34.1 & 71.6 & 87.1 & 110.2 & 143.5 & \textbf{8.3} & \textbf{18.3} & 38.7 & \underline{48.4} & \textbf{65.9} & \underline{102.7} & 10.7 & 25.7 & 60.0 & 76.6 & \underline{106.1} & \underline{164.8} \\
			
			

			\midrule
			M16 & 
			7.5 & 18.7 & 44.8 & 56.4 & 73.6 & 105.2 & 
			13.8 & \underline{31.1} & 66.7 & \underline{80.8} & 102.6 & \textbf{133.9} & 
			8.6 & \underline{18.5} & 39.5 & 49.4 & 67.3 & 103.6 & 
			10.0 & 24.1 & \underline{58.9} & \underline{76.2} & 107.2 & 169.3 \\
			
			M32 &
			7.3 & 18.1 & 43.6 & 55.3 & 72.8 & 105.5 & 
			13.7 & \textbf{31.0} & \underline{65.7} & \textbf{79.9} & \underline{101.4} & \underline{135.7} &
			8.6 & \underline{18.5} & 39.3 & 49.6 & 67.4 & 103.5 &
			\textbf{9.6} & \underline{23.7} & \textbf{57.7} & \textbf{75.0} & \textbf{105.8} & 168.7 \\

			\midrule 
			& 
			\multicolumn{6}{c|}{Purchases} & 
			\multicolumn{6}{c|}{Sitting} &
			\multicolumn{6}{c|}{Sitting Down}&
			\multicolumn{6}{c}{Taking Photo} \\
			
			Time (ms) &
			80 & 160 & 320 & 400 & 560 & 1000 &
			80 & 160 & 320 & 400 & 560 & 1000 &
			80 & 160 & 320 & 400 & 560 & 1000 &
			80 & 160 & 320 & 400 & 560 & 1000 \\
			\midrule 
			
			
			ConvSeq2Seq\cite{Li2018} & 20.3 & 41.8 & 76.5 & 89.9 & 111.3 & 151.5 & 13.5 & 27.0 & 52.0 & 63.1 & 82.4 & 120.7 & 20.7 & 40.6 & 70.4 & 82.7 & 106.5 & 150.3 & 12.7 & 26.0 & 52.1 & 63.6 & 84.4 & 128.1 \\
			
			
			Traj-CNN\cite{Liu2021b} & 14.5 & 31.9 & 66.6 & 80.8 & 103.6 & 141.0 & 11.0 & 21.2 & 45.5 & 57.5 & 79.0 & 120.1 & 16.1 & 29.6 & 58.7 & 72.6 & 97.0 & 147.0 & 10.4 & 20.6 & 44.4 & 55.8 & 76.8 & 120.1 \\
			
			STS-GCN\cite{Sofianos2021b}$^*$ & 13.9 & 31.7 & 66.0 & 80.0 & 102.5 & 142.5 & 9.6 & 20.6 & 45.2 & 57.3 & 79.0 & 122.0 & 15.0 & 29.6 & 59.4 & 73.6 & 98.8 & 149.5 & 9.2 & 19.9 & 43.4 & 55.0 & 76.2 & 118.8 \\
			
			MSR-GCN\cite{Dang2021} & 14.8 & 32.4 & 66.1 & 79.6 & 101.6 & 139.2 & 10.5 & 22.0 & 46.3 & 57.8 & 78.2 & 120.0 & 16.1 & 31.6 & 62.4 & 76.8 & 102.8 & 155.5 & 9.9 & 21.0 & 44.6 & 56.3 & 78.0 & 121.9 \\
			
			MultiAttention\cite{Mao2021} & 18.1 & 36.8 & 58.4 & 67.9 & \textbf{94.5} & \textbf{133.1} & 9.9 & 24.3 & 53.8 & 66.3 & 75.8 & 115.0 & \textbf{10.4} & \textbf{26.6} & \textbf{54.6} & \textbf{66.3} & 96.0 & 141.8 & 5.9 & \textbf{14.8} & 38.0 & 49.4 & \textbf{71.8} & \underline{115.2} \\
			
			DSTD-GCN\cite{Fu2022b} & \underline{12.7} & \underline{29.6} & \underline{62.3} & \underline{75.8} & 97.5 & 137.8 & \textbf{8.8} & \underline{19.3} & 42.9 & 54.3 & 74.9 & 117.8 & 14.1 & 28.0 & \underline{57.3} & 71.2 & 96.1 & 147.2 & \underline{8.4} & 18.8 & 42.0 & 53.5 & 74.5 & 117.9 \\
			
			MotionMixer\cite{Bouazizi2022a} & 14.6 & 31.3 & 62.8 & 76.1 & - & 135.1 & 10.0 & 20.9 & 43.7 & 54.5 & - & 115.7 & \underline{12.0} & 31.4 & 61.4 & 74.5 & - & \underline{141.1} & 9.0 & 18.9 & \underline{41.0} & \textbf{51.6} & - & \textbf{114.6} \\
			
			PGBIG \cite{Ma2022} & \textbf{12.5} & \textbf{28.7} & \textbf{60.1} & \textbf{73.3} & \underline{95.3} & \underline{133.3} & \textbf{8.8} & \textbf{19.2} & \underline{42.4} & \underline{53.8} & 74.4 & 116.1 & 13.9 & \underline{27.9} & 57.4 & 71.5 & 96.7 & 147.8 & \underline{8.4} & 18.9 & 42.0 & 53.3 & 74.3 & 118.6 \\
			
			
			
			\midrule
			M16 & 
			13.0 & 30.3 & 62.8 & 76.7 & 97.9 & 136.2 & 
			\underline{8.9} & 19.4 & 42.7 & 53.9 & \underline{74.2} & \underline{113.4} & 
			14.4 & 30.2 & 58.5 & 71.3 & \underline{95.7} & 141.6 &
			8.5 & 18.6 & \underline{41.0} & \underline{51.7} & \underline{72.9} & 116.4 \\
			
			M32 & 
			13.3 & 30.2 & 63.0 & 77.3 & 97.7 & 134.8 &
			\underline{8.9} & 19.4 & \textbf{42.3} & \textbf{53.6} & \textbf{73.9} & \textbf{113.0} & 
			14.1 & 29.8 & \underline{57.3} & \underline{69.8} & \textbf{94.3} & \textbf{140.2} &
			\textbf{8.2} & \underline{18.4} & \textbf{40.6} & 51.8 & 73.0 & 116.6 \\

			\midrule 
			& 
			\multicolumn{6}{c|}{Waiting} & 
			\multicolumn{6}{c|}{Walking Dog} &
			\multicolumn{6}{c|}{Walking Together}&
			\multicolumn{6}{c}{Average} \\
			
			Time (ms) &
			80 & 160 & 320 & 400 & 560 & 1000 &
			80 & 160 & 320 & 400 & 560 & 1000 &
			80 & 160 & 320 & 400 & 560 & 1000 &
			80 & 160 & 320 & 400 & 560 & 1000 \\
			\midrule 
			
			
			ConvSeq2Seq\cite{Li2018} & 14.6 & 29.7 & 58.1 & 69.7 & 87.3 & 117.7 & 27.7 & 53.6 & 90.7 & 103.3 & 122.4 & 162.4 & 15.3 & 30.4 & 53.1 & 61.2 & 72.0 & 87.4 & 16.6 & 33.5 & 62.0 & 73.5 & 92.1 & 126.8 \\
			
			
			Traj-CNN\cite{Liu2021b} & 10.5 & 21.8 & 45.8 & 56.3 & 73.4 & 104.5 & 21.3 & 43.3 & 80.8 & 94.5 & 115.6 & 153.5 & 10.3 & 21.1 & 38.5 & 44.8 & 54.8 & 68.0 & 12.1 & 24.9 & 50.7 & 62.0 & 80.8 & 114.9 \\
			
			STS-GCN\cite{Sofianos2021b}$^*$ & 10.0 & 21.9 & 47.0 & 58.2 & 76.4 & 107.7 & 20.8 & 43.6 & 81.8 & 95.2 & 114.4 & 151.9 & 10.1 & 20.7 & 39.1 & 46.0 & 54.9 & 62.9 & 11.4 & 24.8 & 51.2 & 62.6 & 81.1 & 113.8 \\
			
			MSR-GCN\cite{Dang2021} & 10.7 & 23.1 & 48.2 & 59.2 & 76.3 & 106.3 & 20.6 & 42.9 & 80.4 & 93.3 & 111.9 & 148.2 & 10.6 & 20.9 & 37.4 & 43.8 & 52.9 & 65.9 & 12.1 & 25.6 & 51.6 & 62.9 & 81.1 & 114.2 \\
			
			MultiAttention\cite{Mao2021} & 9.0 & 22.5 & 55.7 & 71.1 & \underline{72.7} & 105.1 & 29.5 & 54.8 & 100.3 & 105.1 & 119.0 & \underline{141.4} & \textbf{8.0} & \textbf{17.6} & \textbf{33.2} & \underline{42.0} & \underline{51.2} & \underline{63.2} & 11.0 & 23.6 & 49.2 & 60.0 & \textbf{75.9} & \textbf{110.1} \\
			
			DSTD-GCN\cite{Fu2022b} & \underline{8.7} & 20.2 & 44.3 & 55.2 & 73.2 & 105.7 & \underline{19.6} & 41.8 & 77.6 & 90.2 & 109.8 & 147.7 & 9.1 & 19.8 & 36.3 & 42.7 & \textbf{50.5} & \textbf{61.2} & \underline{10.4} & 23.3 & 48.8 & 59.8 & 77.8 & 111.0 \\
			
			MotionMixer\cite{Bouazizi2022a} & 10.2 & 21.1 & 45.2 & 56.4 & - & 107.7 & 20.5 & 42.8 & 75.6 & 87.8 & - & 142.2 & 10.5 & 20.6 & 38.7 & 43.5 & - & 65.4 & 11.0 & 23.6 & \underline{47.8} & 59.3 & - & 111.0 \\
			
			PGBIG \cite{Ma2022} & 8.9 & 20.1 & 43.6 & \textbf{54.3} & \textbf{72.2} & \textbf{103.4} & \textbf{18.8} & \textbf{39.3} & \textbf{73.7} & \underline{86.4} & \underline{104.7} & \textbf{139.8} & \underline{8.7} & \underline{18.6} & \underline{34.4} & \textbf{41.0} & 51.9 & 64.3 & \textbf{10.3} & \textbf{22.7} & \textbf{47.4} & \textbf{58.5} & \underline{76.9} & \underline{110.3} \\
			
			
			
			\midrule
			M16 & 
			8.7 & \underline{19.5} & \underline{43.6} & 54.6 & 73.2 & \underline{104.6} & 
			19.9 & \underline{40.9} & \underline{74.1} & 86.6 & 106.6 & 149.2 & 
			9.7 & 20.4 & 38.5 & 45.8 & 55.2 & 64.7 & 
			10.6 & 23.3 & 48.5 & 59.5 & 77.8 & 110.8 \\   
			
			M32 &  
			\textbf{8.6} & \textbf{19.4} & \textbf{43.5} & \underline{54.8} & 73.6 & 105.4 &
			20.0 & 41.4 & \textbf{73.7} & \textbf{85.1} & \textbf{103.8} & 143.2 &
			9.6 & 20.3 & 38.2 & 45.6 & 55.4 & 64.6 &
			10.5 & \underline{23.2} & 47.9 & \underline{59.0} & 77.2 & \underline{110.3} \\

			\bottomrule
			\hline
		\end{tabular}
		\label{tab:performance_h36m}
	\end{center}
	\vspace{-2mm}
\end{table*}

\begin{table*}[ht]
	\begin{center}
		\tiny
		\setlength{\tabcolsep}{3pt}
		\caption{Performance comparison on different architectures using MPJPE for common action split from the ExPI dataset.}
		\begin{tabular}{c|c@{\hspace{3pt}}c@{\hspace{3pt}}c@{\hspace{3pt}}c@{\hspace{3pt}}c@{\hspace{3pt}}|c@{\hspace{3pt}}c@{\hspace{3pt}}c@{\hspace{3pt}}c@{\hspace{3pt}}c@{\hspace{3pt}}|c@{\hspace{3pt}}c@{\hspace{3pt}}c@{\hspace{3pt}}c@{\hspace{3pt}}c@{\hspace{3pt}}|c@{\hspace{3pt}}c@{\hspace{3pt}}c@{\hspace{3pt}}c@{\hspace{3pt}}c@{\hspace{3pt}}|c@{\hspace{3pt}}c@{\hspace{3pt}}c@{\hspace{3pt}}c@{\hspace{3pt}}c@{\hspace{3pt}}|c@{\hspace{3pt}}c@{\hspace{3pt}}c@{\hspace{3pt}}c@{\hspace{3pt}}c@{\hspace{3pt}}|c@{\hspace{3pt}}c@{\hspace{3pt}}c@{\hspace{3pt}}c@{\hspace{3pt}}c@{\hspace{3pt}}|c@{\hspace{3pt}}c@{\hspace{3pt}}c@{\hspace{3pt}}c@{\hspace{3pt}}c@{\hspace{3pt}}} 
			\toprule 
			& 
			\multicolumn{5}{c|}{A1} & 
			\multicolumn{5}{c|}{A2} &
			\multicolumn{5}{c|}{A3}&
			\multicolumn{5}{c|}{A4} &
			\multicolumn{5}{c|}{A5} & 
			\multicolumn{5}{c|}{A6} &
			\multicolumn{5}{c|}{A7}&
			\multicolumn{5}{c}{AVG} \\
			\midrule 
			Time (ms) &
			0.2 & 0.4 & 0.6 & 0.8 & 1.0 &
			0.2 & 0.4 & 0.6 & 0.8 & 1.0 &
			0.2 & 0.4 & 0.6 & 0.8 & 1.0 &
			0.2 & 0.4 & 0.6 & 0.8 & 1.0 &
			0.2 & 0.4 & 0.6 & 0.8 & 1.0 &
			0.2 & 0.4 & 0.6 & 0.8 & 1.0 &
			0.2 & 0.4 & 0.6 & 0.8 & 1.0 &
			0.2 & 0.4 & 0.6 & 0.8 & 1.0 \\
			
			\midrule 
			
			LTD \cite{Mao2019} & 
			70 & 125 & 157 & - & 189 & 
			131 & 242 & 321 & - & 426 & 
			102 & 194 & 260 & - & 357 & 
			62 & 117 & 155 & - & 197 & 
			72 & 131 & 173 & - & 231 & 
			81 & 151 & 200 & - & 280 & 
			112 & 223 & 315 & - & 442 & 
			90 & 169 & 226 & - & 303 \\
			
			HRI \cite{Mao2020} & 
			52 & 103 & 139 & - & 188 & 
			96 & 186 & 256 & - & 349 &
			57 & 118 & 167 & - & 240 & 
			45 & 93 & 131 & - & 180 &
			51 & 105 & 149 & - & 214 & 
			61 & 125 & 176 & - & 252 &
			71 & 150 & 222 & - & 333 & 
			62 & 126 & 177 & - & 251  \\
			
			MSR-GCN \cite{Dang2021} & 
			56 & 100 & 132 & - & 175 & 
			102 & 187 & 256 & - & 365 &
			65 & 120 & 166 & - & 244 & 
			50 & 95 & 127 & - & 172 &
			54 & 100 & 138 & - & 202 & 
			70 & 132 & 182 & - & 258 &
			82 & 154 & 218 & - & 321 & 
			69 & 127 & 174 & - & 248 \\
			
			XIA \cite{Guo2021} &
			\underline{49} & 98 & 140 & - & 192 &
			\underline{84} & \underline{166} & \underline{234} & - & \underline{346} & 
			\underline{51} & \underline{105} & \underline{154} & - & \underline{234} &
			41 & 84 & 120 & - & \underline{161} & 
			\textbf{43} & \underline{90} & \underline{132} & - & \underline{197} &
			\textbf{55} & \underline{113} & \underline{163} & - & \textbf{242} & 
			\textbf{62} & \textbf{130} & \textbf{192} & - & \textbf{291} &
			\textbf{55} & \underline{112} & \underline{162} & - & \textbf{238} \\
			\midrule 
			
			M16 &
			50 & \underline{92} & \underline{124} & \underline{149} & \underline{173} &
			89 & 171 & 241 & \underline{296} & 350 &
			54 & 109 & 157 & 195 & 235 &
			\underline{40} & \underline{80} & \underline{115} & \underline{141} & 163 &
			47 & 95 & 138 & \underline{174} & 208 &
			\underline{58} & 115 & 166 & 212 & 255 &
			72 & 157 & \underline{236} & \underline{313} & \underline{376} &
			\underline{58} & 117 & 168 & \underline{211} & 251 \\
			
			M32 &
			\textbf{47} & \textbf{89} & \textbf{120} & \textbf{145} & \textbf{169} &
			\textbf{82} & \textbf{159} & \textbf{228} & \textbf{282} & \textbf{338} &
			\textbf{49} & \textbf{100} & \textbf{147} & \textbf{183} & \textbf{218} &
			\textbf{38} & \textbf{76} & \textbf{109} & \textbf{136} & \textbf{159} &
			\underline{45} & \textbf{89} & \textbf{128} & \textbf{162} & \textbf{194} &
			\textbf{55} & \textbf{110} & \textbf{159} & \textbf{203} & \underline{245} &
			\underline{70} & \underline{155} & 238 & \textbf{313} & 378 &
			\textbf{55} & \textbf{111} & \textbf{161} & \textbf{203} & \underline{243} \\
			
			\bottomrule
			\hline
		\end{tabular}
		\label{tab:performance_expi_seen}
	\end{center}
\end{table*}

\begin{table*}[ht]
	\begin{center}
		\tiny
		\vspace{-5mm}
		\setlength{\tabcolsep}{1.5pt}
		\caption{Performance comparison on different architectures using MPJPE for unseen action split from the ExPI dataset.}
		\begin{tabular}{c|c@{\hspace{1.5pt}}c@{\hspace{1.5pt}}c@{\hspace{1.5pt}}c@{\hspace{1.5pt}}c@{\hspace{1.5pt}}|c@{\hspace{1.5pt}}c@{\hspace{1.5pt}}c@{\hspace{1.5pt}}c@{\hspace{1.5pt}}c@{\hspace{1.5pt}}|c@{\hspace{1.5pt}}c@{\hspace{1.5pt}}c@{\hspace{1.5pt}}c@{\hspace{1.5pt}}c@{\hspace{1.5pt}}|c@{\hspace{1.5pt}}c@{\hspace{1.5pt}}c@{\hspace{1.5pt}}c@{\hspace{1.5pt}}c@{\hspace{1.5pt}}|c@{\hspace{1.5pt}}c@{\hspace{1.5pt}}c@{\hspace{1.5pt}}c@{\hspace{1.5pt}}c@{\hspace{1.5pt}}|c@{\hspace{1.5pt}}c@{\hspace{1.5pt}}c@{\hspace{1.5pt}}c@{\hspace{1.5pt}}c@{\hspace{1.5pt}}|c@{\hspace{1.5pt}}c@{\hspace{1.5pt}}c@{\hspace{1.5pt}}c@{\hspace{1.5pt}}c@{\hspace{1.5pt}}|c@{\hspace{1.5pt}}c@{\hspace{1.5pt}}c@{\hspace{1.5pt}}c@{\hspace{1.5pt}}c@{\hspace{1.5pt}}|c@{\hspace{1.5pt}}c@{\hspace{1.5pt}}c@{\hspace{1.5pt}}c@{\hspace{1.5pt}}c@{\hspace{1.5pt}}|c@{\hspace{1.5pt}}c@{\hspace{1.5pt}}c@{\hspace{1.5pt}}c@{\hspace{1.5pt}}c@{\hspace{1.5pt}}} 
			\toprule
			& 
			\multicolumn{5}{c|}{A8} &
			\multicolumn{5}{c|}{A9} &
			\multicolumn{5}{c|}{A10}&
			\multicolumn{5}{c|}{A11}&
			\multicolumn{5}{c|}{A12}& 
			\multicolumn{5}{c|}{A13}&
			\multicolumn{5}{c|}{A14}&
			\multicolumn{5}{c|}{A15}&
			\multicolumn{5}{c|}{A16}&
			\multicolumn{5}{c}{AVG} \\
			\midrule 
			Time (ms) &
			0.2 & 0.4 & 0.6 & 0.8 & 1.0 &
			0.2 & 0.4 & 0.6 & 0.8 & 1.0 &
			0.2 & 0.4 & 0.6 & 0.8 & 1.0 &
			0.2 & 0.4 & 0.6 & 0.8 & 1.0 &
			0.2 & 0.4 & 0.6 & 0.8 & 1.0 &
			0.2 & 0.4 & 0.6 & 0.8 & 1.0 &
			0.2 & 0.4 & 0.6 & 0.8 & 1.0 &
			0.2 & 0.4 & 0.6 & 0.8 & 1.0 &
			0.2 & 0.4 & 0.6 & 0.8 & 1.0 &
			0.2 & 0.4 & 0.6 & 0.8 & 1.0 \\
			
			\midrule 
			
			LTD \cite{Mao2019} &
			- & 239 & 324 & 394 & - &
			- & 175 & 226 & 259 & - &
			- & 148 & 191 & 220 & - &
			- & 176 & 240 & 286 & - &
			- & 143 & 178 & 192 & - &
			- & 146 & 193 & 226 & - &
			- & 252 & 333 & 387 & - &
			- & 174 & 228 & 264 & - &
			- & 139 & 184 & 217 & - &
			- & 177 & 233 & 272 & - \\
			
			HRI \cite{Mao2020} & 
			- & 195 & 283 & 358 & - &
			- & 121 & 169 & 206 & - &
			- & 92 & 129 & 160 & - &
			- & 129 & 193 & 245 & - &
			- & 80 & 104 & 121 & - &
			- & 112 & 154 & 187 & - &
			- & 157 & 219 & 257 & - &
			- & 134 & 190 & 233 & - &
			- & 96 & 146 & 187 & - &
			- & 124 & 176 & 218 & - \\
			
			MSR-GCN \cite{Dang2021} &
			- & 230 & 289 & 335 & - &
			- & 188 & 245 & 290 & - &
			- & 148 & 198 & 248 & - &
			- & 234 & 319 & 384 & - &
			- & 176 & 232 & 278 & - &
			- & 162 & 218 & 266 & - &
			- & 177 & 239 & 295 & - &
			- & 143 & 179 & 213 & - &
			- & 157 & 222 & 281 & - &
			- & 179 & 238 & 288 & - \\
			
			XIA \cite{Guo2021} & 
			- & 191 & 287 & 377 & - &
			- & 118 & 165 & 203 & - &
			- & \textbf{91} & \textbf{129} & \textbf{162} & - &
			- & \textbf{122} & \textbf{183} & \textbf{232} & - &
			- & \textbf{81} & \textbf{107} & \textbf{128} & - &
			- & 106 & \underline{150} & 185 & - &
			- & \textbf{156} & \textbf{216} & \textbf{256} & - &
			- & \underline{126} & 175 & 213 & - &
			- & \underline{96} & 152 & 205 & - &
			- & \underline{121} & 174 & 218 & - \\
			\midrule 

			M16 &
			\underline{54} & \underline{115} & 170 & 220 & \underline{257} &
			\underline{55} & \underline{90} & \underline{110} & \underline{128} & \underline{153} &
			\underline{52} & \underline{104} & \underline{147} & \underline{185} & \underline{219} &
			\underline{80} & 156 & 213 & 256 & \underline{293} &
			\underline{65} & 130 & 183 & 226 & \underline{260} &
			\underline{43} & \underline{88} & \textbf{129} & \underline{171} & \underline{212} &
			\underline{94} & 192 & 282 & 356 & \underline{410} &
			\underline{63} & \underline{126} & \underline{172} & \underline{210} & \underline{258} &
			\underline{50} & 98 & \underline{135} & \underline{166} & \underline{203} &
			\underline{62} & 122 & \underline{171} & \underline{213} & \underline{252 }\\
			
			M32 &
			\textbf{50} & \textbf{112} & 169 & 219 & \textbf{257} &
			\textbf{51} & \textbf{86} & \textbf{105} & \textbf{121} & \textbf{145} &
			\textbf{52} & 107 & 151 & 190 & \textbf{225} &
			\textbf{77} & \underline{150} & \underline{203} & \underline{243} & \textbf{278} &
			\textbf{61} & \underline{123} & \underline{174} & \underline{215} & \textbf{246} &
			\textbf{43} & \textbf{87} & \textbf{129} & \textbf{169} & \textbf{210} &
			\textbf{87} & \underline{185} & \underline{277} & \underline{354} & \textbf{413} &
			\textbf{58} & \textbf{120} & \textbf{166} & \textbf{208} & \textbf{252} & 
			\textbf{46} & \textbf{89} & \textbf{120} & \textbf{148} & \textbf{187} & 
			\textbf{58} & \textbf{118} & \textbf{166} & \textbf{207} & \textbf{246} \\
			\bottomrule
			\hline
		\end{tabular}
		\label{tab:performance_expi_unseen}
	\end{center}
	\vspace{-2mm}
\end{table*}

\begin{table*}[ht]
	\caption{(left) (a) Performance comparison between different architectures using MPJPE for the AMASS and 3DPW datasets. ($^*$) metric is computed by us using our pipeline and the standard metric. ($^\dag$) metric takes the average MPJPE over all previous frames. (right) (b) Comparison summary of average MPJPE (using only 80, 160, 320, 400, and 1000ms), number of parameters and FLOPs.}
	\begin{subtable}[ht]{0.65\linewidth}
		\vspace{-1mm}
		\centering
		\notsotiny
		\begin{tabular}{c|c@{\hspace{5pt}}c@{\hspace{5pt}}c@{\hspace{5pt}}c@{\hspace{5pt}}c@{\hspace{5pt}}c@{\hspace{5pt}}c@{\hspace{5pt}}c@{\hspace{5pt}}|c@{\hspace{5pt}}c@{\hspace{5pt}}c@{\hspace{5pt}}c@{\hspace{5pt}}c@{\hspace{5pt}}c@{\hspace{5pt}}c@{\hspace{5pt}}c@{\hspace{5pt}}}
			\toprule 
			& 
			\multicolumn{8}{c|}{AMASS-BMLrub} & 
			\multicolumn{8}{c}{3DPW} \\
			Time (ms) &
			80 & 160 & 320 & 400 & 560 & 720 & 880 & 1000 &
			80 & 160 & 320 & 400 & 560 & 720 & 880 & 1000 \\
			\midrule 
			
			GAGCN\cite{Zhong2022}$^\dag$ & 
			10.0 & 11.9 & 20.1 & 24.0 & 30.4 & - & - & 43.1 & 8.4 & 11.9 & 18.7 & 23.6 & 29.1 & - & - & 39.9 \\
			
			HRI\cite{Mao2020} & 
			11.3 & 20.7 & 35.7 & 42.0 & 51.7 & 58.6 & 63.4 & 67.2 & 12.6 & 23.1 & 39.0 & 45.4 & 56.0 & 63.6 & 69.7 & 73.7 \\
			STS-GCN\cite{Sofianos2021b}$^*$ & 
			11.2 & 20.6 & 36.5 & 43.1 & 52.5 & 59.2 & 64.3 & 68.7 & 11.7 & 20.7 & 35.0 & 40.3 & 48.7 & 55.0 & 59.4 & 62.4 \\
			MultiAttention\cite{Mao2021} & 
			11.0 & 20.3 & 35.0 & 41.2 & 50.7 & 57.4 & 61.9 & 65.8 & 12.4 & 22.6 & 38.1 & 44.4 & 54.7 & 62.1 & 67.9 & 71.8 \\
			MotionMixer\cite{Bouazizi2022a}$^*$ &
			10.1 & \textbf{18.4} & \textbf{32.7} & \textbf{38.9} & \textbf{48.3} & \textbf{55.0} & \underline{60.4} & \underline{64.2} & 
			10.9 & \underline{19.4} & \underline{33.3} & \underline{39.0} & 48.4 & 55.2 & 60.0 & 63.6 \\
			
			\midrule 
			M16 & \underline{9.9} & 18.9 & 34.1 & 40.4 & 50.2 & 56.9 & 61.3 & 64.9 & \underline{10.6} & 19.6 & 33.4 & \underline{39.0} & \underline{48.0} & \underline{54.1} & \underline{58.8} & \underline{62.0} \\
			
			M32 & \textbf{9.8} & \underline{18.6} & \underline{33.6} & \underline{39.8} & \underline{49.2} & \underline{56.0} & \textbf{60.3} & \textbf{63.6} & \textbf{10.4} & \textbf{19.3} & \textbf{33.2} & \textbf{38.7} & \textbf{47.6} & \textbf{54.0} & \textbf{58.5} & \textbf{61.7} \\
			
			\bottomrule
			\hline
		\end{tabular}
		\phantomcaption
		\label{tab:performance_amas_3dpw}
	\end{subtable}
	\hfill
	\begin{subtable}[ht]{0.32\linewidth}
		\centering
		\notsotiny
		\begin{tabular}{c|c@{\hspace{5pt}}c@{\hspace{5pt}}c@{\hspace{5pt}}}
			\toprule 
			& \multicolumn{3}{c}{Human3.6M} \\
			Model & MPJPE & Params & $\approx$FLOPs \\
			\midrule 
			MultiAttention\cite{Mao2021} & 50.8 & 3.42M & 142.3M \\
			MSR-GCN\cite{Dang2021} & 53.3 & 6.3M & 192.4M \\
			STS-GCN\cite{Sofianos2021b} & 52.8 & 57.5k & 7.1M \\
			MotionMixer\cite{Bouazizi2022a} & 50.5 & \textbf{30.2K} & \textbf{2.1M} \\
			DSTD-GCN\cite{Fu2022b} & 50.6 & 0.18M & - \\
			PGBIG \cite{Ma2022} & \underline{49.8} & 1.74M & 55.8M \\
			\midrule 
			M8 & 50.8 & 115.6K & 19.5M \\
			M16 & 50.5 & 164.0K & 21.3M \\
			M32 & 50.2 & 345.6K & 27.5M \\
			M64 & \textbf{49.6} & 1.048M & 49.7M \\
			\bottomrule
			\hline
		\end{tabular}
		\phantomcaption
		\label{tab:complexity}
	\end{subtable}
	\vspace{-2.5mm}
\end{table*}

\subsection{Model architecture}
Motivated by the interpretability of the feature importance of random forest, we built a model not only for pose sequence prediction but also for output understanding via feature importance and connectivity matrices similar to previous works \cite{Sofianos2021b, Fu2022b, Zhong2022}. This architecture is shown in Fig. \ref{fig:overview}. We argue that our results generate feature maps that can be used to observe and figure out unexpected behaviors in certain OOD data. More concretely, Fig. \ref{fig:architecture} uses an encoder-decoder architecture but splits the temporal and spatial GCNs. Inspired by DeepLabv3+ \cite{Chen2018}, we propose to replace the original TCN described in \cite{Sofianos2021b} with a new Atrous Pyramid TCN (APTCN). Also, we propose the Context Network (ConNet) and the Dynamic Spatio-Temporal Graph Convolutions Network (DST-GCN) placed in parallel that sum the results with a global residual connection to obtain the final pose sequence. Additionally, following previous works that use trajectory representation successfully as inputs \cite{Su2022, Liu2021a, LiuYong2019}. We propose to use an overall of 10 input dimensions, 3 for joint positions, 6 for joint instant velocities and accelerations in x,y and z, and 1 $L_2$-norm vector of the instant velocities. We support the idea that the last two layers from the model, DST-GCN and ConNet, can satisfactorily obtain interpretability from the output pose sequence whereas the DST-GCN blocks in the input can extract relevant information regarding the input sequence.

\textbf{Interpretable-GCN layer.}
We implement DST-GCN split into the Dynamic Spatial Graph Network (DSGN) and Dynamic Temporal Graph Network (DTGN) that are controlled by a Gating Network (GaNet) as shown in \ref{fig:DSTGCN}, $F$ is set given the model size and is detailed in Section \ref{sec:experiment}. The interpretable information from DST-GCN is located in the adjacency matrices of the GCNs. In contrast to most GCNs, we consider learning sample-specific connections in the adjacency matrix has a more meaningful representation and also helps to interpret graph connections. To do this, we replace the adjacency matrix with a Dynamic Adjacency Encoder (DAE) that provides a feature map of the same size, described below.

\textbf{Atrous Pyramid TCN.}
TCN architecture is widely used as a decoder for the output sequences \cite{Sofianos2021b, Fu2022b, Zhong2022}. However, as explained above, we use a larger input dimension and can increase the complexity of the feature search in the output sequence. Given the outstanding results obtained by DeepLabv3+ in image segmentation, we modified the TCN to be pyramidal and used different dilation rates to later concatenate and compress the output, as shown in Fig. \ref{fig:APTCN}.

\textbf{Context Network.}
We propose a network to collect statistics and generate feature importance vectors, detailed in Fig. \ref{fig:Context}. We think every pooling extracts different feature information, as observed in point clouds \cite{Ionescu2014, LiuYong2019}. Specifically, we use 3 different pooling operations for the same input. Conv+BN+PReLU blocks and linear layers are applied before and after each pooling. Later, every output with size $o \in \mathbb{R}^{T}$ is merged in one unique vector with size $O \in \mathbb{R}^{3T}$. Assuming we code the context information, we could extract two feature importance from this vector for spatial and temporal domains: displacement and joint features.

\textbf{Gating Network (GaNet).}
After obtaining the output from DSGN and DTGN, we presume that not all feature maps contribute equally, similar to \cite{Zhong2022}, we weight the feature maps. But our GaNet blocks generate two vectors $W1$ and $W2$ with $F$ length but with a different fusion mode. GaNet, as shown in Fig. \ref{fig:Gating}, implements a lightweight architecture with the use of separable convolutions that reduce the number of parameters. Also, the aggregation of statistic values computed from the same input makes the representation more meaningful to the weighting vectors.

\textbf{Dynamic Adjacency Encoder (DAE).}
This block, as shown in Fig. \ref{fig:DAE}, is responsible to compute the adjacency matrix used in every GCN. This efficient architecture uses convolutional layers not only to map an input feature map into a matrix adjacency shape but also to generate this matrix with a lower number of parameters than other GCNs.

\begin{figure*}
	\centering
	\begin{subfigure}[b]{0.40\linewidth}
		\centering
		\includegraphics[width=\linewidth]{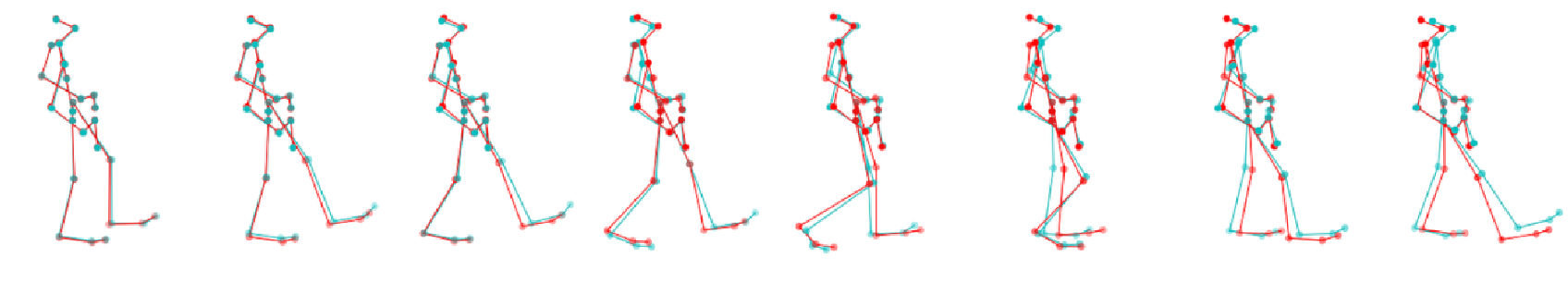}
	\end{subfigure}
	\hspace{0.05\textwidth}
	\begin{subfigure}[b]{0.40\linewidth}
		\centering
		\includegraphics[width=\linewidth]{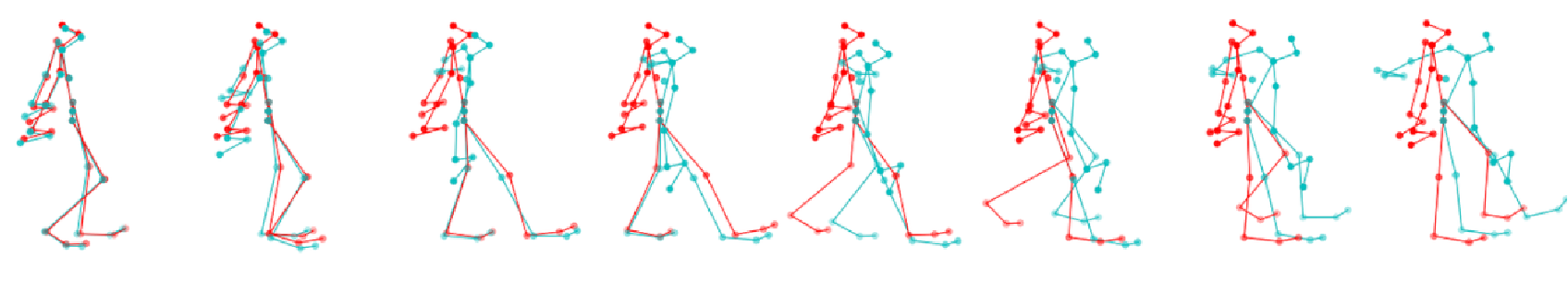}
	\end{subfigure}
	\begin{subfigure}[b]{0.40\linewidth}
		\centering
		\includegraphics[width=\linewidth]{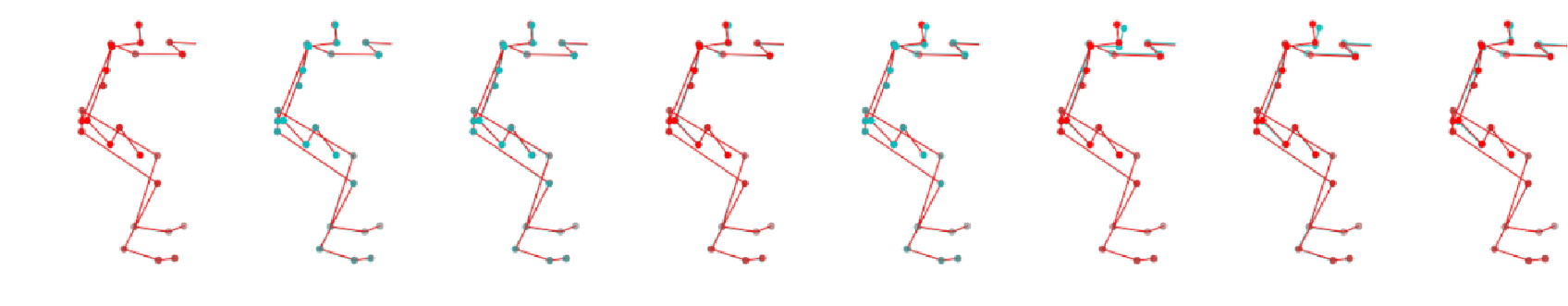}
	\end{subfigure}
	\hspace{0.05\textwidth}
	\begin{subfigure}[b]{0.40\linewidth}
		\centering
		\includegraphics[width=\linewidth]{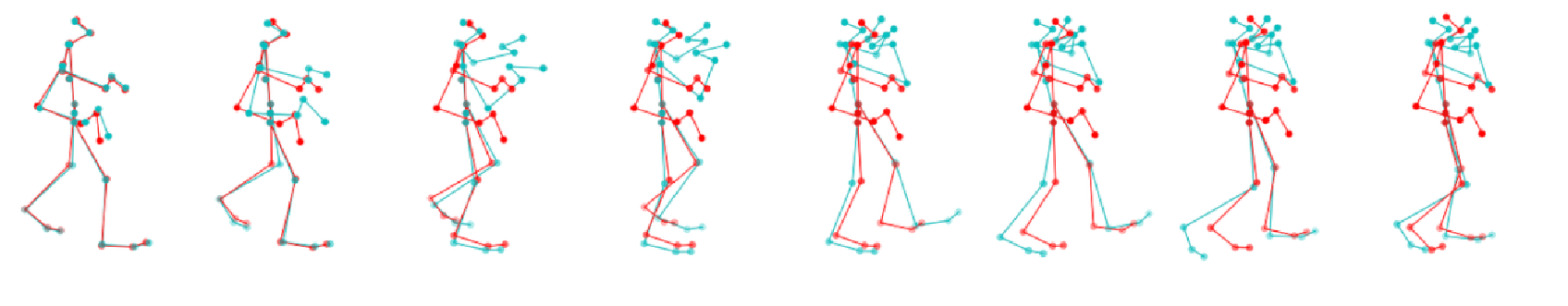}
	\end{subfigure}
	\vspace{-5mm}
	\caption{Motion prediction results on ``walking" (top) and ``eating" (bottom) motion classes from H3.6M dataset. Sorted by the lowest (left) and the largest errors (right). Solid lines are ground truth. Dashed lines are predictions from the $M32$ model. Blue color of the poses represents ground truth while the red color of the poses represents the predicted ones.}
	\label{fig:qualitative}
\end{figure*}

\section{Experimental Evaluation}
\label{sec:exper}

\subsection{Datasets}
Using a unique model to evaluate short- and long-predictions, same as \cite{Mao2020, Mao2021}, we conducted experiments on widely used datasets: Human 3.6M \cite{Ionescu2014}, AMASS \cite{Mahmood2019}, and 3DPW \cite{VonMarcard2018}. Additionally, we analyzed multi-pose motion prediction using the ExPI dataset \cite{Guo2021}.

\textbf{Human 3.6M}. Consists of 15 different actions performed by 7 different actors per action. Following \cite{Mao2020} and \cite{Fu2022b}, we downsample the frame rate to 25Hz and use 22 joints from the overall subjects 1,6,7,8,9 for training, subject 11 for validation, and subject 5 for testing.

\textbf{AMASS}. Consists of a gathering of 18 existing datasets. We perform a frame rate down-sampling to 25Hz as in Human 3.6M. Then, following \cite{Sofianos2021b} and \cite{Bouazizi2022a}, we select 8, 4, and 1 (BMLrub) datasets for training, validation, and testing respectively. For each body pose, hand joints are discarded and we consider 18 joints for training from the 22 body joints, skipping 5 frames instead of 1.

\textbf{3DPW}. Consists of both indoor and outdoor actions, containing 51,000 frames captured at 30Hz. Following \cite{Sofianos2021b} and \cite{Bouazizi2022a}, we only use 3DPW to test the generalization of the models trained on AMASS.

\textbf{ExPI}. The recent ExPI dataset contains 115 sequences of 2 professional couples performing 16 different dance actions. It is recorded in a multiview motion capture studio at 25 fps. Following \cite{Guo2021}, The experiments performed include two persons represented by 18 joints each one in all the 30K frames. We benchmark in both protocols for common and unseen action splits using the same settings.

\subsection{Experimental results}
\label{sec:experiment}

\textbf{Metric}. The Mean Per Joint Position Error (MPJPE) is a widely adopted evaluation metric used in previous works \cite{Lyu2022d, Dang2021, Liu2021b} to compare two pose sequences and is described in Eq. \ref{eq:mpjpe}. We evaluate our topology defined as M and the number of channels in the hidden DST-GCN layers for various experiments, i.e. $M8$, $M16$, $M32$, and $M64$.

\begin{equation}
	\mathcal{L}_{\text{MPJPE}} = \frac{1}{J \times T} \sum^{T}_{t=1}\sum^{J}_{j=1}{ \lVert \hat{x}_{j,t} - x_{j,t} \rVert _2}
	\label{eq:mpjpe}
\end{equation}

\begin{figure}
	\centering
	\begin{subfigure}[b]{0.42\linewidth}
		\centering
		\includegraphics[width=\linewidth]{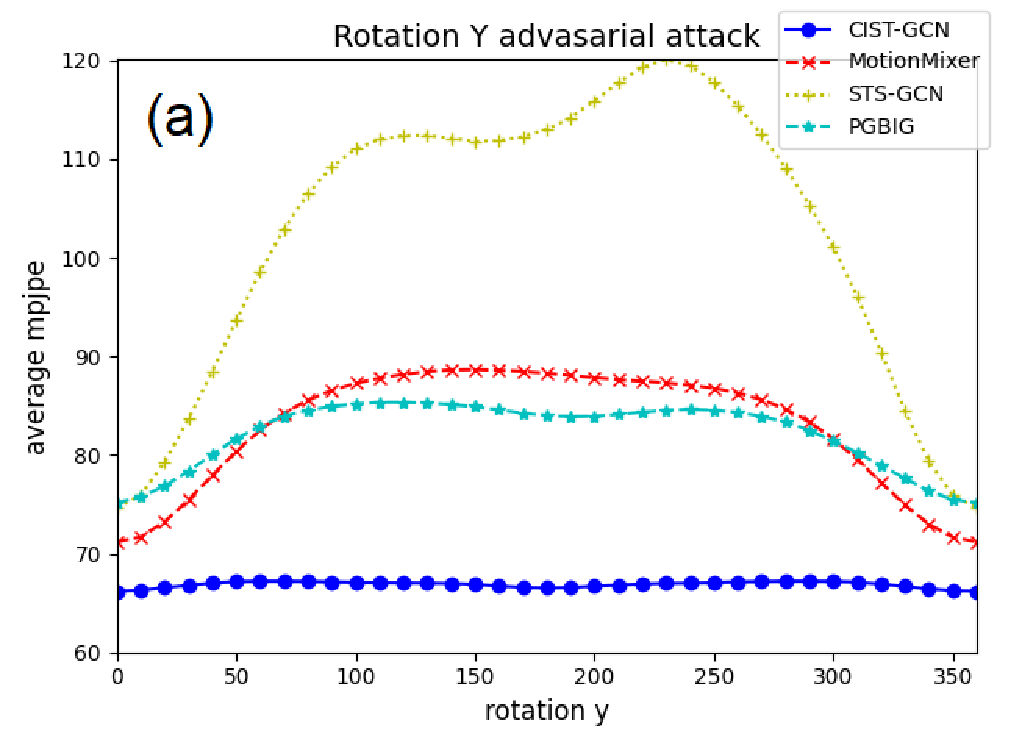}
		\phantomcaption
		\label{fig:randomrotationy}
	\end{subfigure}
	\vspace{-4mm}
	\begin{subfigure}[b]{0.42\linewidth}
		\centering
		\includegraphics[width=\linewidth]{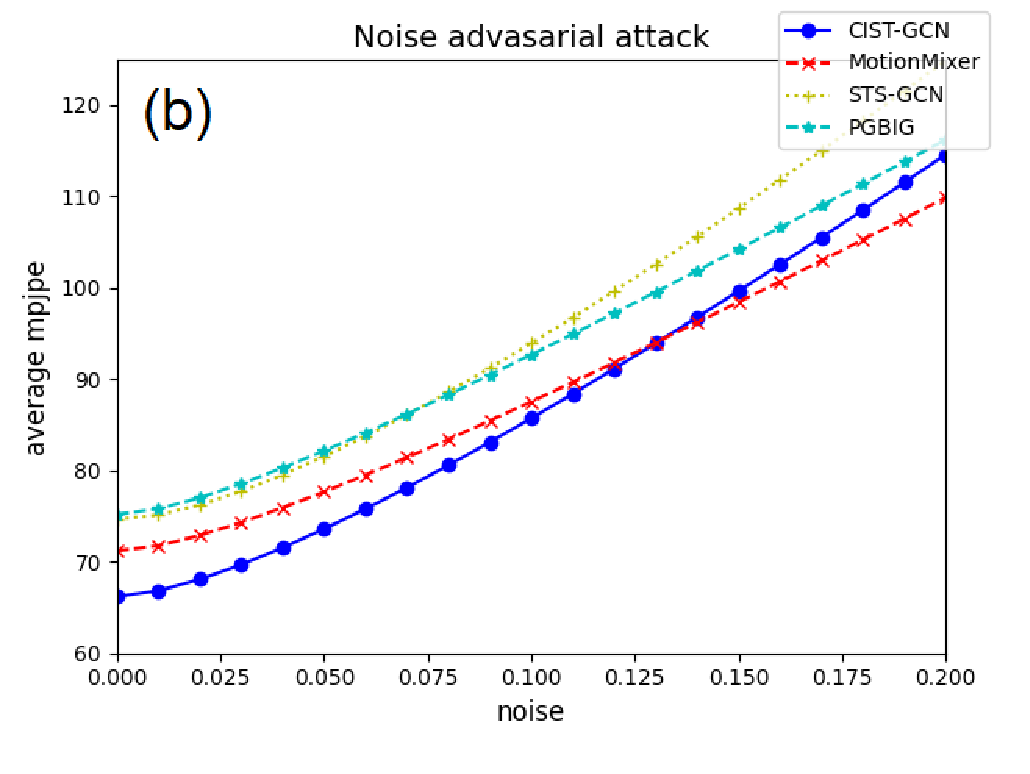}
		\phantomcaption
		\label{fig:randomnoise}
	\end{subfigure}
	\vspace{-4mm}
	\caption{Augmentation effect on test set evaluated on our pipeline. Average MPJPE over the 25 output frames, with (a) rotations between 0-360 degrees, and (b) noise rate between 0.0-0.2.}
	\label{fig:augmentation}
	\vspace{-4mm}
\end{figure}

\textbf{Quantitative results}.
We first compare our method with the SOTA approaches on the four datasets. Since we realized that some papers used another metric, which calculates the mean error across preceding frames with diverse normalizations, resulting in a lower error compared to the standard metric. we attempted to reproduce these outcomes whenever feasible implementations were obtainable. but our replication efforts relied solely on the standard metric defined in Eq. \ref{eq:mpjpe}. While running these experiments, we found other noteworthy peculiarities in the literature: Firstly, some authors did not use all motion classes for comparison. Secondly, some authors sampled 256 samples from each motion class while others used the complete test sets. We chose to use the 256-sample variation, as it is more common in the literature and made more sense for a balanced test set. We present in Tab. \ref{tab:performance_h36m} our model results compared to the benchmark architectures for H3.6M. Regrettably, we did not find an implementation for GAGCN \cite{Zhong2022}. In general, we observe that more complex actions such as “Purchase”, “Sitting Down”, “Posing” and “Walking Dog” yield lower performances for all methods since these “spontaneous movements” have large motion variations appear after the input sequence. Although we outperformed most of the SOTA models, we also obtained slightly lower but still comparable performance to MultiAttention and PGBIG models. But, these models are larger in both the number of parameters and demanding complexity as detailed in section \ref{sec:complexity}. The newest MotionMixer \cite{Bouazizi2022a} architecture obtains comparable results using only a feature mix while keeping a lower number of parameters. We also experimented with OOD samples by introducing 3D transformations and noise as adversarial examples to evaluate their impact on the models. Our model demonstrates its robustness on the test set against random rotation attacks (Fig. \ref{fig:randomrotationy}) when compared to other models, and, performs slightly better than other models, for random noise attacks (depicted in Fig. \ref{fig:randomnoise}). This shows the importance of data augmentation in the overall robustness of the model against “natural perturbations”.

We also extend our experiments to the AMASS and 3DPW datasets as detailed in Tab. \ref{tab:performance_amas_3dpw}. Our $M32$ and $M16$ models outperform previous works and $M16$ is comparable to MotionMixer. On AMASS, we observe that for some models, the error is similar for short-term predictions while differing more for long-term predictions. Interestingly, MotionMixer reduced significantly the short-term error but the long-term margin remains similar. We obtain a similar behavior to MotionMixer with our two models and assume that this is because of the nature of AMASS dataset which contains many complex samples that are not necessarily cyclic such as “Walking” action from H3.6M. On 3DPW dataset, we observe that long-term predictions differ more between models, and still $M32$ and $M16$ outperform previous approaches. We believe that this behavior happens due to spontaneous movements, even when we find some “Walking” motion in both datasets. GAGCN is not directly comparable to the other methods shown in the table.

Finally, we wanted to explore the multi-pose motion interaction in the new ExPI dataset and how CIST-GCN may behave in this situation. CIST-GCN was initially designed for single-pose predictions. However, it impressively reaches previous SOTA methods in both short-term and long-term. In Tabs. \ref{tab:performance_expi_seen} and \ref{tab:performance_expi_unseen}, our model outperforms other approaches by a short margin independent of the protocol type. But, our model complexity is only 0.76M on $M32$ compared to 8.5M from XIA model \cite{Guo2021}.

\begin{figure}
	\centering
	\begin{subfigure}{0.42\linewidth}
		\includegraphics[width=\linewidth]{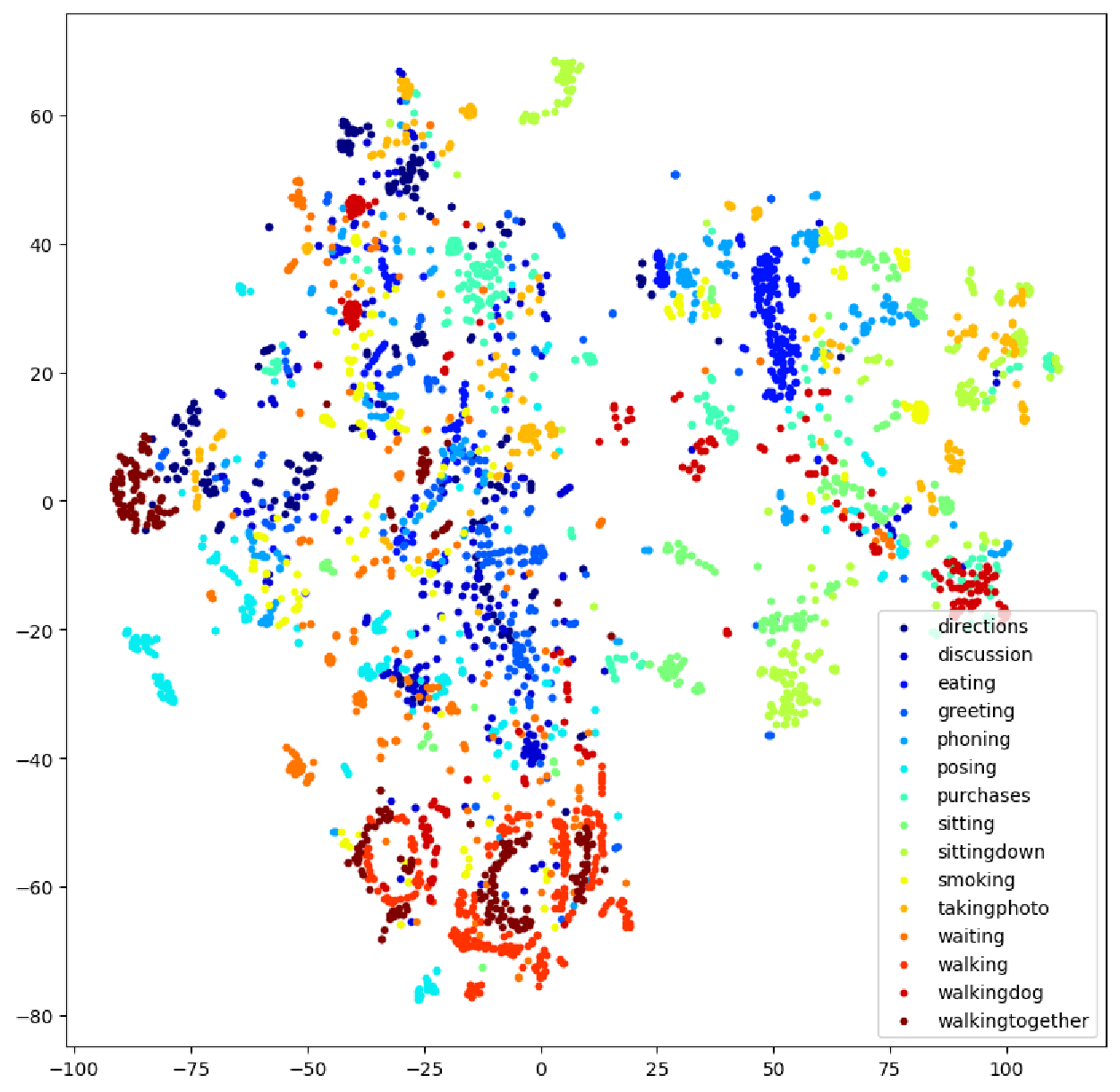}
		\phantomcaption
		\label{fig:in_3D}
	\end{subfigure}
	\vspace{-4mm}
	\begin{subfigure}{0.42\linewidth}
		\includegraphics[width=\linewidth]{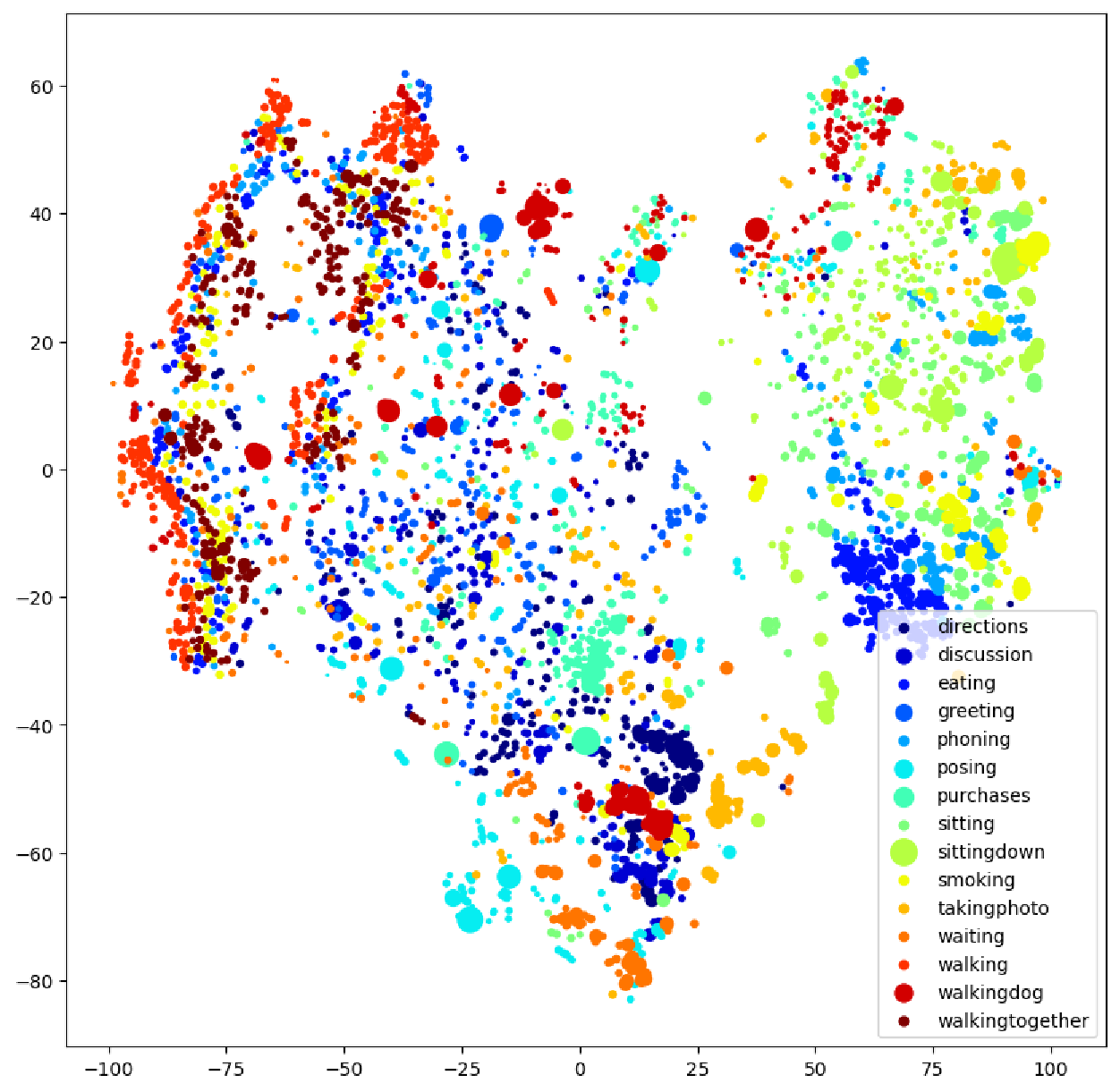}
		\phantomcaption
		\label{fig:feature_importance}
	\end{subfigure}
	\vspace{-4mm}
	\caption{t-sne representation of the test set using (a) input poses (b) all feature importance from the model concatenated. MPJPE values are represented by scatter size.}
	\label{fig:features_visualization}
\end{figure}

\textbf{Qualitative results}. In order to complement the quantitative results, we present in Fig. \ref{fig:qualitative} the results of the $M32$ model for input sequences of the ``walking" and ``eating" classes from the H3.6M test set. The plot shows the samples with the lowest (left) and largest (right) errors with predictions of 80, 160, 320, 400, 560, 720, and 1000 ms. As we can see, the right side sample (OOD or most difficult case) is arguably very hard to predict without context knowledge, for example, “walking and raising your hand after the input period”. This might even be considered as unexpected behavior after the input sequence which is similar to an OOD.

\subsection{Computational Complexity}
\label{sec:complexity}
We assess the trade-off between the performance and the approximated computational cost of SOTA architectures versus our model across 4 complexity configurations, as presented in Table \ref{tab:complexity}. We use the average MPJPE value as a performance reference next to a number of parameters and FLOPs approximation. We observe that our architecture has a lower number of parameters compared to most of the previous works and outperforms other approaches while being lightweight. However, due to the interpretable features implemented in our model, the number of FLOPs increased significantly due to the matrix multiplications and linear layers. Although $M64$ obtained the lowest error and overcomes previous works, it requires a large number of parameters and sometimes has convergence issues. We believe that the sum operator in the DST-GCN blocks can sometimes generate overflow or large gradients, making training unstable. Therefore, we focus on $M32$ for further analysis because of the accuracy-complexity trade-off. PGBIG and MotionMixer have the best MPJPE values from the SOTA models, however, the estimated complexity (FLOPs) for PGBIG is larger due to it requires a multi-stage architecture with intermediate targets whereas MotionMixer underperforms in terms of MPJPE by a small margin our model.

\subsection{Implementation details}
Our model was trained end-to-end and in a fully supervised manner using Eq. \ref{eq:mpjpe} as the loss function for all the experiments. We used data augmentation composed of random noises, rotations, scales, and translations. Inspired by scalable modeling, we control the model size by two hyper-parameters, complexity, and the number of layers. We stack the DST-GCN module five times for the input but only apply the DST-GCN module once for the output layer as shown in Fig. \ref{fig:architecture}. The complexity was set on 8, 16, 32, and 64 for simplicity (details in supplementary material and our code).

\begin{figure*}
	\centering
	\begin{subfigure}{0.40\linewidth}
		\includegraphics[width=\linewidth]{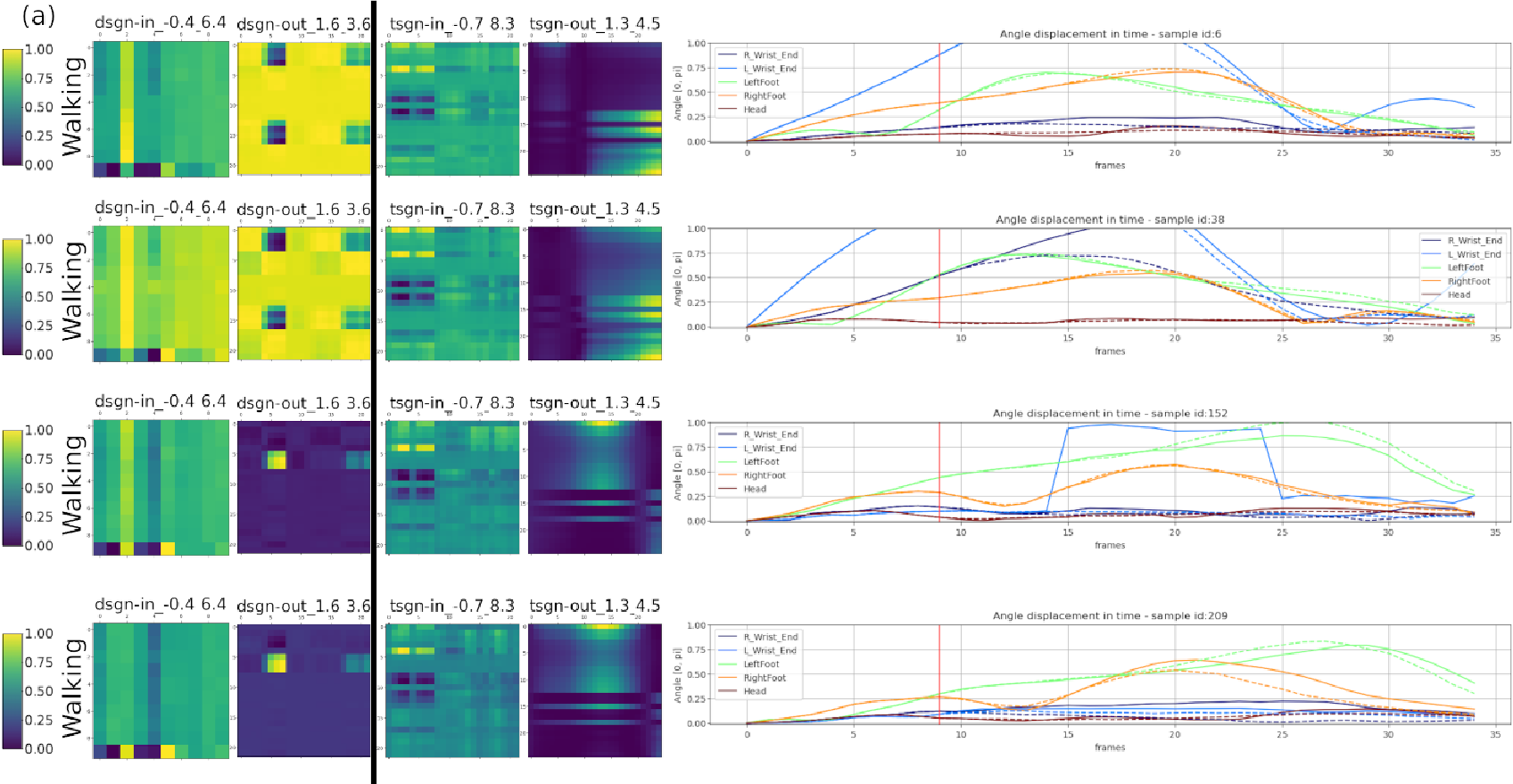}
		\phantomcaption
		\label{fig:feature_maps_walking}
	\end{subfigure}
	\vspace{-4mm}
	\hspace{0.05\textwidth}
	\begin{subfigure}{0.40\linewidth}
		\includegraphics[width=\linewidth]{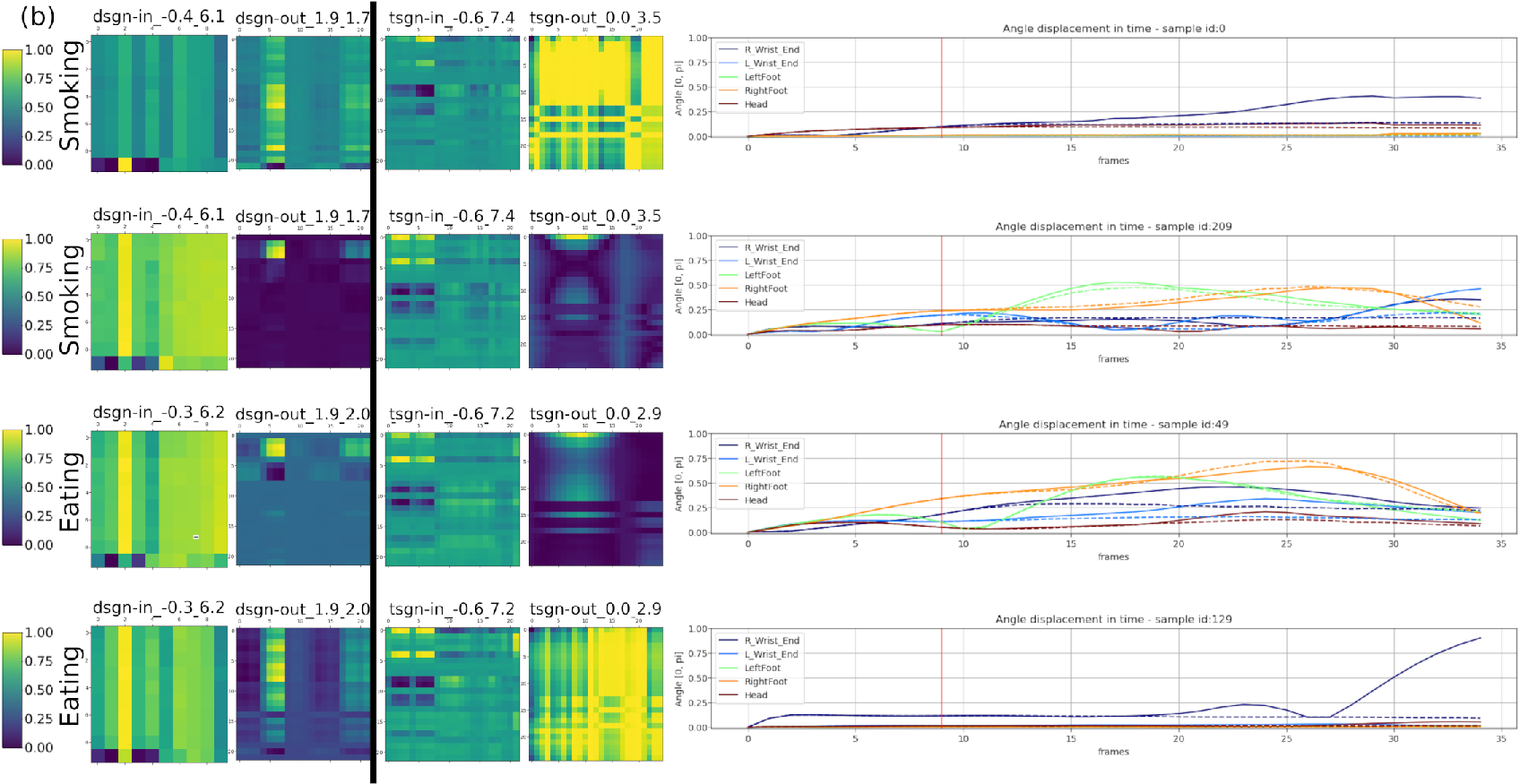}
		\phantomcaption
		\label{fig:feature_maps_others}
	\end{subfigure}
	\vspace{-4mm}
	\caption{Normalized (0-1) and per-layer average adjacency matrices extracted from the CIST-GCN architecture in the spatial (left) and temporal (right) domains for (a) walking, and (b) other motion actions. The right parts display changes in the angles of movement.}
	\label{fig:feature_maps_visualization}
\end{figure*}

\begin{figure}
	\centering
	\includegraphics[width=\linewidth]{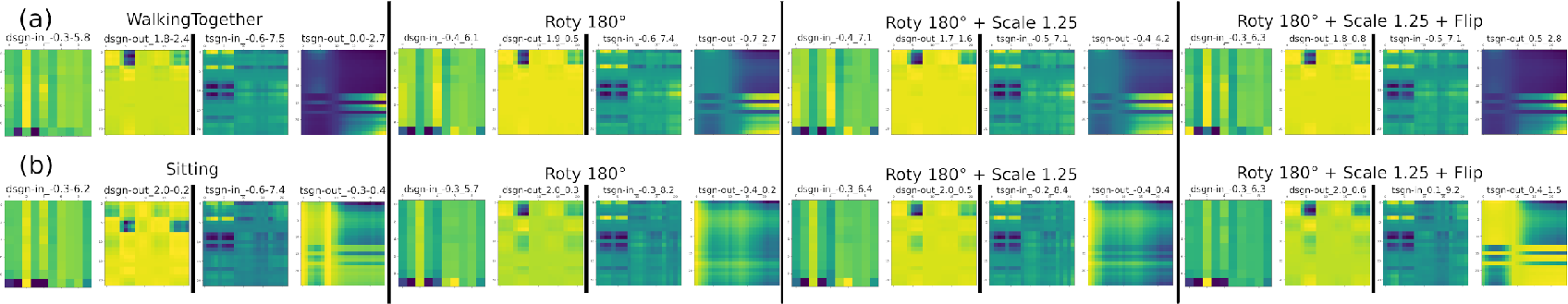}
	\caption{Normalized (0-1) and per-layer average adjacency matrices for (a) ``walkingtogether" and (b) ``sitting" motion actions.}
	\label{fig:feature_maps_attack}
	\vspace{-6mm}
\end{figure}

\section{Discussion}
\label{sec:discussion}

In this section, we focus on the $M32$ model. Additional details can also be found in the supplementary materials.

\subsection{Feature importance vectors}
We explore the significance of interpretations learned from the model by comparing them to another data representation, as depicted in Fig. \ref{fig:features_visualization}. We concatenate all features’ importance obtained by every model layer. In contrast to the approach in GAGCN \cite{Zhong2022}, where authors computed the average of 16 samples from 4 motion classes and plotted differences of the blending coefficients, we use the t-SNE algorithm to visualize the entire test set. This approach avoids interpretation bias, especially when certain classes share similar movements, as shown in Fig. \ref{fig:qualitative}. In Fig. \ref{fig:features_visualization}, only pure 3D euclidean poses and corresponding feature importance representations are shown (see supplementary material for displacement representations). Our observation reveals that using pure input poses, as seen in Fig. \ref{fig:in_3D}, results in a cluster-like distribution visualization for some motion sequences, while others exhibit less pronounced grouping and higher variance without distinct centroids. In comparison to other representations, Fig. \ref{fig:feature_importance} effectively shows well-grouped motion classes, while also unveiling instances of larger MPJPE located away from the centroids. This experiment found a similar grouping as in GAGCN \cite{Zhong2022} for ``walking" and ``sitting". A similar interpretation could be seen in other representations using other grouping strategies. Quantitatively, we measured distances from points to centroids and found that our data representation was more accurate. We can also utilize average vectors for both the temporal and spatial domains to conduct soft clustering on movements. Also, given our architecture uses a global residual connection, the model indeed is learning displacements from the last input. This is particularly useful in inference since movements can be ambiguous sometimes, leading to instances where a sequence may exhibit a blend of multiple motion actions such as ``walking" and ``eating". Then, we could predict the future motion by also getting an idea about the kind of motion that is observed.

\subsection{Feature Maps}
After evaluating Fig. \ref{fig:feature_importance}, we observe that motion classes may behave as a mix of at least two motion classes. A qualitative analysis of the saliency maps shows us the motion behavior. Consistent with DSTD-GCN \cite{Fu2022b}, the interpretable adjacency matrices are equivalent in some way to relation matrices. In Fig. \ref{fig:feature_maps_walking}, we present the saliency maps, along with the spatial and temporal matrices, corresponding to walking actions, as well as the associated variations in movement angles (see supplementary materials for relative angle computation). We observe that when the input sequence comprises cyclic movements like walking, the temporal saliency maps (``tsgn-out" for the output) prominently feature values close to 0. On the other hand, the output spatial saliency maps (``dsgn-out") present lower values when the right foot starts the cyclic movement before the left foot, we observe the opposite behavior when these spatial maps are mostly 1. We have observed that when the input sequence comprises cyclic movements like walking, the output temporal saliency maps (``tsgn-out") prominently feature values close to 0. Conversely, the output spatial saliency maps (dsgn-out) exhibit lower values when the cyclic movement begins with the right foot before the left. By contrast, when the left foot begins the movement, the spatial maps predominantly display values near to 1. To broaden this analysis to additional motion classes, we observe in Fig. \ref{fig:feature_maps_others} that similar patterns persist in ``tsgn-out" when the movement is cyclic, however, when the model predict static motions ``tsgn-out" displays a prevalence of values near 1. Conversely, when we evaluate the averages of the input saliency maps (``tsgn-in" and ``dsgn-in"), the patterns are faded in the hidden layers, making motion identification a challenging task. Sometimes, the saliency maps exhibit inconsistencies for the highest MPJPE values. The saliency maps should show the relation between frames and joints, however, it becomes apparent that the precise values are distributed across consecutive rows and columns. This phenomenon is evident when we present in Fig. \ref{fig:feature_maps_visualization} the saliency label alongside the map's mean and standard deviation. Similar patterns are observed when input sequences are rotated or slightly scaled as shown in Fig. \ref{fig:feature_maps_attack}. Essentially, this shows that the matrices exhibit similarity for similar predictions, enabling us to derive comparable interpretations irrespective of the object's position, orientation, or scale. Given our model forecasts displacements, the output layers contain more substantial and visually prominent information. Additional experiments are necessary to have a deeper comprehension of which layer(s) within our network transform input displacements into output displacements.

\section{Conclusions and Future Outlook}
\label{sec:conclusion}
We have introduced a novel architecture for human motion prediction using GCNs. The evaluations show that our approach obtains comparable and/or superior performance to SOTA models. As observed, our model is a robust approach not only for motion prediction but also for achieving a certain interpretability level. We discussed the effects of data augmentation and OOD data, also showed the robustness of our models against previous works. For future outlook, we plan to extend our study on adversarial attacks and OOD to have a deeper understanding of the feature maps and how the model complexity can be optimized.

\section*{Acknowledgment}
The research leading to these results is funded by the German Federal Ministry for Economic Affairs and Climate Action within the project “ATTENTION – Artificial Intelligence for realtime injury prediction”. The authors would like to thank the consortium for the successful cooperation.

{\small
	\bibliographystyle{ieee_fullname}
	\bibliography{PaperForReview}{}	
}

\end{document}